\documentclass[10pt,journal,compsoc]{IEEEtran}
\ifCLASSOPTIONcompsoc
  \usepackage[nocompress]{cite}
\else
  \usepackage{cite}
\fi
\usepackage{algorithm}
\usepackage{algorithmicx}
\usepackage[noend]{algpseudocode}
\usepackage{bm}
\usepackage{amsmath}
\usepackage{amssymb}
\usepackage{amsthm}
\usepackage[pdftex]{graphicx}
\usepackage{balance}
\usepackage{array}
\usepackage{enumitem}
\graphicspath{{figs/}}
\algrenewcommand\algorithmicrequire{\textbf{Input:}}
\algrenewcommand\algorithmicensure{\textbf{Output:}}

\usepackage{pdfpages}

\usepackage[capitalise]{cleveref}

\usepackage{xcolor}

\def\BibTeX{{\rm B\kern-.05em{\sc i\kern-.025em b}\kern-.08emT\kern-.1667em\lower.7ex\hbox{E}\kern-.125emX}}

\newcommand{\transpose}{^{\ensuremath{\mathsf{T}}}}
\newtheorem{theorem}{\textbf{Theorem}}

\newcommand{\Real}{\mathbb{R}}

\newcommand{\CC}{\mathcal{C}}  
\newcommand{\CH}{\mathcal{H}}
\newcommand{\CG}{\mathcal{G}}  
\newcommand{\CL}{\mathcal{L}}  
\newcommand{\CN}{\mathcal{N}}  
\newcommand{\RCN}{\mathring{\mathcal{N}}}
\newcommand{\bA}{\boldsymbol{A}}  

\newcommand{\bH}{\boldsymbol{H}}  
\newcommand{\bX}{\boldsymbol{X}}  
\newcommand{\bZ}{\boldsymbol{Z}}  
\newcommand{\bI}{\boldsymbol{I}}  
\newcommand{\bD}{\boldsymbol{D}}

\newcommand{\bS}{\boldsymbol{S}}  

\newcommand{\bh}{\boldsymbol{h}}  
\newcommand{\bx}{\boldsymbol{x}}  
\newcommand{\by}{\boldsymbol{y}}  
\newcommand{\bW}{\boldsymbol{W}}  
\newcommand{\half}{\frac{1}{2}}
\newcommand{\layer}[1]{^{(#1)}}
\DeclareMathOperator*{\argmin}{argmin}
\algnewcommand{\LineComment}[1]{\State \(\triangleright\) #1}

\newcommand{\header}[1]{\noindent\textbf{#1}}
\newcommand{\bullethdr}[1]{\smallskip\noindent\textbullet\,\textbf{#1}}

\begin{document}

\title{Federated Learning over Coupled Graphs}

%

\author{Runze~Lei,
        Pinghui~Wang,
        Junzhou~Zhao,
        Lin~Lan,
        Jing~Tao,
        Chao~Deng,
        Junlan~Feng,~\IEEEmembership{Fellow,~IEEE},
        Xidian~Wang,
        Xiaohong~Guan,~\IEEEmembership{Fellow,~IEEE} 

\IEEEcompsocitemizethanks{
\IEEEcompsocthanksitem
R. Lei, P. Wang, J. Zhao, J. Tao, and L. Lan are with the MOE Key Laboratory for Intelligent Networks and Network Security,
Xi'an Jiaotong University, P.O. Box 1088,
No. 28, Xianning West Road, Xi'an, Shaanxi 710049, China.
E-mail: xjtu2140506016@stu.xjtu.edu.cn, \{phwang, junzhou.zhao, jtao\}@xjtu.edu.cn, llan@sei.xjtu.edu.cn.
\IEEEcompsocthanksitem C. Deng and J. Feng is with China Mobile Research Institute.
E-mail: \{dengchao, fengjunlan\}@chinamobile.com.
\IEEEcompsocthanksitem X. Wang is with China Mobile Group Design Institute.
E-mail: westpointwp@qq.com.
\IEEEcompsocthanksitem X. Guan is with the MOE Key Laboratory for Intelligent Networks and
Network Security, Xi'an Jiaotong University, P.O. Box 1088, No. 28,
Xianning West Road, Xi'an, Shaanxi 710049, China 
and also
with the Center for Intelligent and Networked Systems, Tsinghua National
Lab for Information Science and Technology, Tsinghua University, Beijing
100084, China. E-mail: xhguan@mail.xjtu.edu.cn.
}
\thanks{Manuscript received August 3, 2022; revised November 18, 2022.}
\thanks{(Corresponding author: Pinghui Wang.)}
\thanks{Digital Object Identifier no. }
}

\markboth{Journal of \LaTeX\ Class Files,~Vol.~14, No.~8, August~2015}%
{Shell \MakeLowercase{\textit{et al.}}: Bare Demo of IEEEtran.cls for Computer Society Journals}

\IEEEtitleabstractindextext{%
\begin{abstract}
Graphs are widely used to represent the relations among entities.
When one owns the complete data, an entire graph can be easily built, therefore performing analysis on the graph is straightforward.
However, in many scenarios, it is impractical to centralize the data due to data privacy concerns.
An organization or party only keeps a part of the whole graph data, i.e., graph data is isolated from different parties.
Recently, Federated Learning (FL) has been proposed to solve the data isolation issue, mainly for Euclidean data.
It is still a challenge to apply FL on graph data because graphs contain topological information which is notorious for its non-IID nature and is hard to partition.
In this work, we propose a novel FL framework for graph data, FedCog, to efficiently handle coupled graphs that are a kind of distributed graph data, but widely exist in a variety of real-world applications such as mobile carriers' communication networks and banks' transaction networks.
We theoretically prove the correctness and security of FedCog.
Experimental results demonstrate that our method FedCog significantly outperforms traditional FL methods on graphs.
Remarkably, our FedCog improves the accuracy of node classification tasks by up to $14.7\%$.
\end{abstract}

\begin{IEEEkeywords}
Graph neural networks, federated learning, privacy preservation, distributed graph processing.
\end{IEEEkeywords}}

%
%
%
\maketitle

\section{Introduction} \label{sec:introduction}

\IEEEPARstart{G}{RAPHS} are pervasive to represent real-world data such as social relations,
traffic networks, financial transactions, and knowledge databases.
To build a graph, we collect the items of interest as nodes and the relations
among them as edges.
Ideally, all nodes and edges in a graph are owned by a single party.
Hence, the whole graph can be easily built and performing graph
analysis on the graph is straightforward.

However, in many real-world scenarios, different parts of the graph data
are often owned by different parties.
Due to benefit conflicts, security issues, legal restrictions, privacy concerns,
etc., each party's graph data should be protected and is forbidden to be
accessed by other parties.
For example, a mobile carrier (resp. a bank) keeps the information of its
customers, along with the communication (resp. financial transaction) records among customers.
Each mobile carrier's data forms an individual graph, which is highly confidential.
Meanwhile, customers in one mobile carrier (resp. a bank) may have connections to customers in another
mobile carrier (resp. another bank), e.g., an AT\&T customer may call a T-Mobile customer.
Thus, each party's graph is only a subgraph of the underlying
graph, which is distributed on multiple parties, and a party cannot
access other parts of the graph possessed by other parties.

\begin{figure}[tbp]
	\centering
	\vspace{-15pt}
	\includegraphics[width=0.93\linewidth]{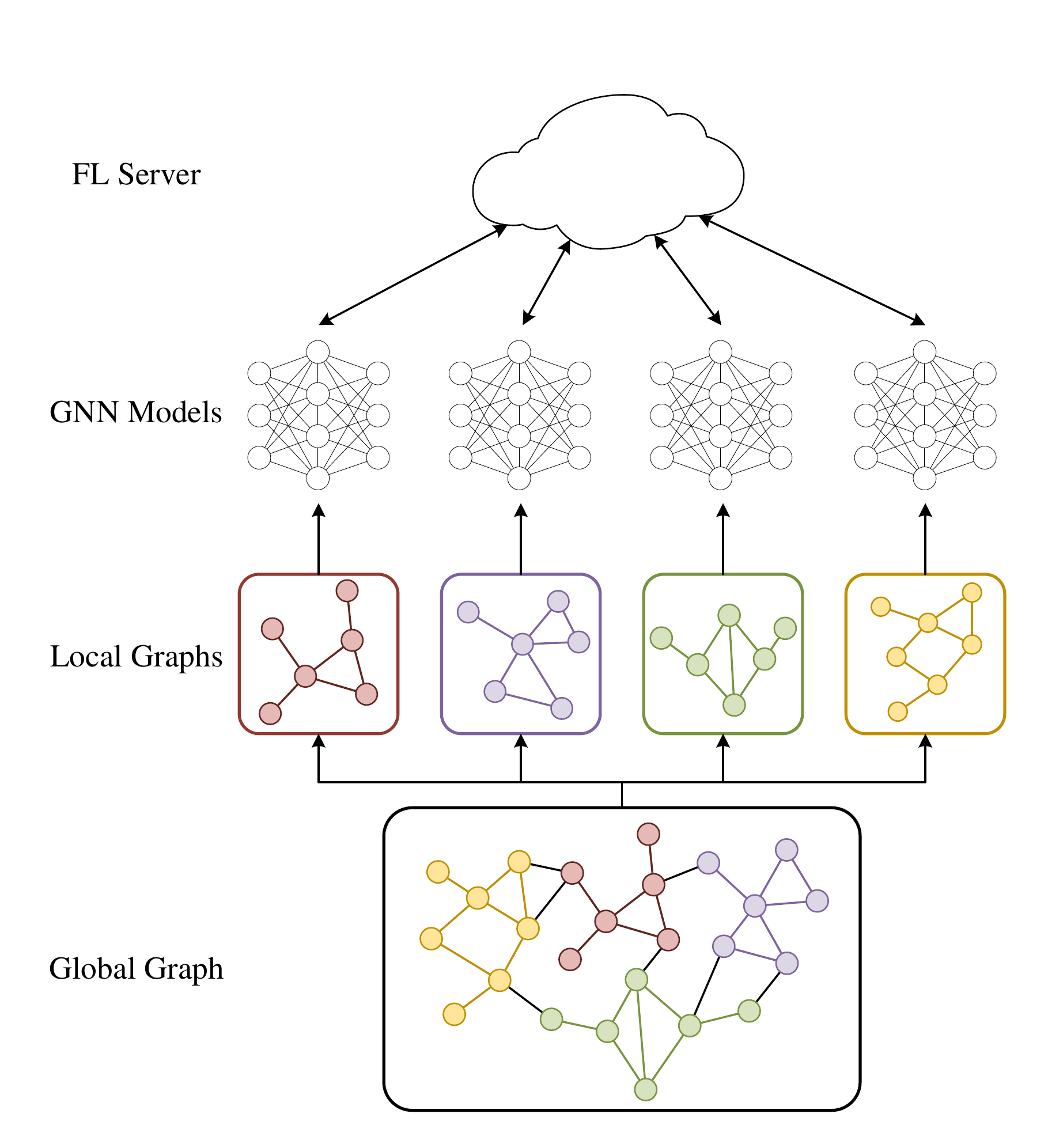}
	\vspace{-10pt}
	\caption{The general structure of federated learning on distributed graph data. Each FL party possesses a subgraph of the global graph. However, a portion of topological information on the global graph is missing on the local graphs. Some edges on the global graph connect nodes held by different FL parties, so none of the parties can utilize these edges when training the GNN models.}
	\label{fig:subgraph_fl}
\end{figure}

In the literature, Federated Learning (FL)~\cite{McMahanMRHA17, YangLCT19} is a
machine learning framework to handle the data isolation issue,
which collaboratively unites a set of parties to jointly train a model while preserving each party's data privacy.
Recently, FL has also been applied to handle distributed graph data~\cite{MeiGLP19,ZhengZCWWZ21,abs-2104-07145}.
The general structure of federated learning on distributed graph data is shown in~\cref{fig:subgraph_fl}.
Each FL party maintains a subgraph of the global graph and trains GNN models on its local graph data.
Most prior works (e.g., FedGraphNN~\cite{abs-2104-07145}) treat the graph data located on each party independently
and ignore inter-connections between different parties.
In other words, these graph FL methods do not concern the \emph{coupling} of parties' graph data.
However, the coupling of graph structure widely exists.
For example, different mobile carriers' communication networks are coupled by
inter-phone calls (i.e., calls between users of different carriers), and different banks' transaction networks are coupled by
inter-money transactions.
These \emph{inter-edges} are critical for some applications (e.g., money laundering
detection), while it is challenging to exploit these edges in this FL setting.

FedSage+~\cite{ZhangYLSY21} is proposed as a subgraph FL method based on graph generation.
Generative models are trained by graph impairing and then generate the missing neighbor nodes and edges.
Thus, FL parties can train GNN models on the mended local graphs and improve the performance of federated GNN models.
However, the generated nodes tend to follow the local distribution, instead of representing the missing neighbors on the other subgraphs.
FedSage+ resolves this problem by introducing a loss term that computes the distances between generated node features and external features.
This scheme increases the similarities between generated local nodes and external nodes, but is also sensitive to the initial values of the generative model and suffers from node mismatching.
Therefore, FedSage+ may not work properly in highly non-IID FL tasks.

In this work, we propose a novel FL framework for graph data, \emph{FedCog},
to efficiently handle {\em coupled graphs}, where the inter-edges between different
parties are critical for graph learning tasks.
In the setting of FedCog, graph data is distributed on several parties, and
each party keeps a set of nodes as well as their features.
A party's graph data not only consists of intra-edges that connect its local
nodes but also inter-edges that connect its local nodes to the nodes possessed by
other parties.
We theoretically prove the correctness and security of FedCog.
We show that performing Graph Convolutional Network (GCN)~\cite{KipfW17} on distributed graph data with FedCog is the same as performing GCN over the underlying global graph.

Our main contributions are summarized as follows.
\begin{itemize}[leftmargin=3ex]
	\item We formally define the problem of FL over coupled graphs.
	The problem is general and widely exists in many real-world applications.
	(\cref{sec:problem})
	
	\item We propose FedCog, an effective method to solve the problem of FL over
	coupled graphs.
	The method can be easily integrated with many existing FL frameworks and popular GNN
	models.
	(\cref{sec:method})
	
	\item We conduct extensive experiments on 12 real-world datasets.
	Compared with  straightforward FL methods, FedCog outperforms by up to $14.7\%$ accuracy in node classification tasks and $0.232$ AUC in link prediction tasks.
	(\cref{sec:results})
\end{itemize}

\section{Preliminaries}
\label{sec:preliminaries}
\header{Graph Convolutional Networks} (GCNs)~\cite{KipfW17} approximate spectral graph convolutions to build GNN models. 
Formally, let $\CG=(V,E,\bX)$ denote the graph of interest, where $V=\{v_1,\ldots,v_n\}$ is a set
of nodes, $E$ is a set of edges, and $\bX\in\Real^{n\times F}$ is the node feature
matrix.
Let $\bA\in\Real^{n\times n}$ denote the adjacency matrix of $\CG$, and the $(u,v)$ entry $a_{uv}$ indicates whether there exists an edge from node $u$ to node $v$.
We define the degree matrix $\bD=\mathrm{diag}(\{d_u\}_{u\in V})$ where $d_u =
\sum_{v\in V}a_{uv}$ is the degree of node $u$.
Let $\tilde{\bA}\triangleq \bA + \bI$ denote the Self-Loop-Augmented (SLA) adjacency matrix~\cite{arenas2008analysis}, which is obtained by adding an identity matrix to the original adjacency matrix of the graph.
Besides, $\tilde\bD$ is the degree matrix of $\tilde\bA$.
Define $\bS$ as the normalized adjacency matrix, i.e.,
\begin{equation}\label{eq:S}
	\bS\triangleq\tilde{\bD}^{-\frac{1}{2}}\tilde{\bA}\tilde{\bD}^{-\frac{1}{2}}.
\end{equation}

GCN computes node embeddings via layer-wise neighborhood aggregation and representation
transformation, and learns node embeddings by stacking multiple layers.
Specifically, GCN computes nodes' $l$-th layer embedding as:
\begin{equation}
	\bH\layer{l+1}=\sigma(\bS\bH\layer{l}\bW\layer{l}), \quad l=0,1,\ldots,L-1,
\end{equation}
where $\sigma(\cdot)$ is a non-linear activation function, $\bH^{(l)}$ and $\bW\layer{l}$ are the node embedding matrix and the trainable feature transformation parameters at the $l$-th layer, respectively.
$L$ is the number of GCN layers.
GCN sets $\bH^{(0)}=\bX$.
It is easy to find that the above graph convolution can also be represented as a feature propagation from each node to its neighbors, that is,
\begin{equation}\label{eq:gcn_prop}
	\bh_u\layer{l+1}=\sigma\left(\sum_{v\in \CN_u}c_{uv}\bh_v\layer{l}\bW\layer{l}\right), \quad u\in V,
\end{equation}
where $\bh_u\layer{l}$ is the representation of node $u$,
$\CN_u=\{v: (u,v)\in E\}\cup\{u\}$ is is the set of node $u$'s neighbors in the graph $\CG$ with SLA adjacency matrix $\tilde{\bA}$, 
and $c_{uv}=1/\sqrt{(1+d_u)(1+d_v)}$.

\header{Simple Graph Convolution} (SGC)~\cite{WuSZFYW19} hypothesizes that the
non-linearity between GCN layers is not critical and can be removed.
For the $l$-th GCN layer, SGC converts the input node representation matrix
$\bH^{(l)}$ to output node representation matrix $\bH^{(l+1)}$ by
\begin{equation}\label{eq:HL}
	\bH^{(l+1)} = \bS\bH^{(l)}.
\end{equation}
SGC iteratively operates Eq.~\eqref{eq:HL} for $L$ layers, followed by a
softmax operation in the final layer to obtain probabilistic outputs.
SGC with such simplification tricks achieves comparable performance as the regular GCN in many applications while greatly outperforming in efficiency.

\header{Federated Learning} aims to learn model parameters $\theta$ that minimizes the loss over all parties,
where parties jointly learn a model while preserving each party's data privacy.
If $m$ parties participate in the FL task, and each party holds local data and labels $(\bX_i, \by_i)$, the optimization task is
\begin{equation}\label{eq:fed_def}
	\begin{gathered}
		\min_{\theta} F^{\text{fed}}(\theta)= \sum_{i=1}^{m}p_i F_i(\theta),\\
		F_i(\theta)=\CL(\operatorname{NN}(\bX_i;\theta),\by_i),
	\end{gathered}
\end{equation}
where $p_i$ is the weight for party $i$'s local objective function.
$\operatorname{NN}$ indicates a neural network model with parameter $\theta$.
$\CL$ is the loss function.
FedAvg~\cite{McMahanMRHA17} uses federated gradient descent to solve the
optimization problem.
FedAvg computes local gradients $g_i=\nabla_\theta F_i(\theta)$
on each party $i$, and aggregates the local gradients by weighted averaging
$\bar{g} = \sum_i p_i g_i$.
The average gradient is then used to update the model parameters $\theta$ via
gradient descent.

\section{Formulated Problem}\label{sec:problem}

\subsection{Coupled Graphs}\label{sec:couple_problem}

In this work, we consider a situation that a graph $\CG=(V, E, X)$ is distributed over $m$
parties and each party can only access its own data.
These parties want to jointly learn a model to address graph learning tasks
such as node classification and link prediction.
Specifically, the graph data accessible by party $i$ is denoted by $\CG_i=(V_i,
V_i^*, E_i, \bX_i)$,
where $V_i\subseteq V$ is the set of {\em internal nodes} of which features $\bX_i$ are possessed by party $i$ 
and $V_i^*$ is a set of {\em external border nodes} that party $i$'s internal nodes connect to, i.e., $V_i^*=\{v^*: (v, v^*)\in E, v\in V_i, v^*\notin V_i\}$.
A node's feature vector is generally private data and is only held by one party.
Thus, in this paper, we assume that different parties' internal nodes have no overlapping,
i.e., $V_i \cap V_j= \emptyset$, $i\ne j$.
$E_i$ is the set of edges accessible by party $i$, including both {\em intra-edges} (i.e., edges between internal nodes) and {\em inter-edges} (i.e.,
edges from internal nodes to external nodes).
We refer to subgraphs $\{\CG_i\}_{i=1}^{m}$ as {\em coupled graphs}.

Take the mobile phone network as an instance.
$\CG_i$ refers to mobile carrier $i$'s communication network.
$V_i$ represents the set of the mobile carrier's customers, and $\bX_i$ records
its customers' features (e.g., age and gender).
Thus, $\bX_i$ is private information of party $i$ and should never be revealed to the server or other parties during training. 
$V_i^*$ consists of other carriers' customers that mobile carrier $i$'s customers communicated with.
For any customer in $V_i^*$, its features are \emph{not} available for mobile carrier $i$.
The edges represent communications between customers.
The external nodes and inter-edges are essential information for the mobile carriers' service.
Therefore, external nodes and inter-edges in the local graph $\CG_i$ do not cause privacy risk.
Many real-world scenarios are similar, such as financial networks between banks, document networks between academic databases, and social networks between email service providers.
These scenarios can be modeled as coupled graph problems and solved by our method.

\subsection{Federated Learning on Coupled Graphs}

We aim to perform graph learning tasks such as node/graph classification and link prediction on coupled graphs $\{\CG_i\}_{i=1}^m$ under the constraint that party $i$ can only access its own graph data $\CG_i$.
In regular FL tasks, local losses $\CL_i, i=1,\ldots, m$ are independent, i.e., $\CL_i$ and its gradient $g_i$ could be locally computed on party $i$.
Thus, FedAvg~\cite{McMahanMRHA17} achieves gradient descent for a global loss $\CL$ by aggregating local gradients.
For graph learning tasks, information from neighbor nodes is crucial.
If we suppose that the server collects subgraph data without concern for privacy, the ideal global model is
\begin{equation}\label{eq:fed_central}
	\theta^*=\argmin_{\theta}\CL(\operatorname{GNN}(E,\bX;\theta),\by),
\end{equation}
where $E$ is graph $\CG$'s edge set, and $\bX, \by$ are node features in graph $\CG$ and target labels respectively.
The GNN model can utilize complete neighbor information of graph $\CG$ in this case.
However, access to the nodes and edges of other parties is restricted due to data privacy.
The neighborhood and long-range relationship of nodes are damaged when ignoring these external nodes and edges.
If the subgraphs are disconnected during federated GNN training, the local optimization task in Eq.~\eqref{eq:fed_def} becomes
\begin{equation}\label{eq:fed_disconnected}
	F_i(\theta)=\CL(\operatorname{GNN}(E_i^{\text{local}},\bX_i;\theta),\by_i),
\end{equation}
where $E_i^{\text{local}}=E_i\cap (V_i\times V_i)$ is the local edge set of subgraph $\CG_i$.
We observe that the federated objective shown by Eq.~\eqref{eq:fed_def} and \eqref{eq:fed_disconnected} may be distinctly different from that in Eq.~\eqref{eq:fed_central} because a part of neighbor information is missing in the federated scheme.
Our experiments later in \cref{sec:results} reveal that
the disconnection of coupled graphs will significantly deteriorate the performance of FL models.

In this work, we propose FedCog which constructs the coupled information on parties, still without direct access to sensitive data.
Our method encourages the parties to share desensitized information securely.
The insensitive data $\boldsymbol{Z}$ are utilized to compose valid knowledge for other FL parties.
The local task of FedCog is
\begin{equation}\label{eq:obj_coupled}
	F_i^{\text{coupled}}(\theta)=\CL(\operatorname{FedCog}(E_i,\bX_i, \{\bZ_{j\to i}\}; \theta), \by_i),
\end{equation}
where $\bZ_{j\to i}$ represents external data from party $j$ to $i$.
Our FedCog allows foreign data to engage in local training and addresses the missing neighbor problem.
Meanwhile, data privacy is theoretically guaranteed by our method.

\section{Our Method}\label{sec:method}

As we mentioned,
the primary difficulty of the coupled graph problem is the data dependence among parties.
A party's GCN model needs representations of all neighboring nodes to achieve graph convolution (as shown in Eq.~\eqref{eq:gcn_prop}).
However, in the coupled graph problem, the neighboring nodes are possessed by other data owners who refuse to share their original graph data.
Our method FedCog provides a surrogate approach to reduce data dependence.
FedCog uses a novel graph transform operation {\em Graph Decoupling}.
Graph Decoupling converts each original local graph into two decoupled local graphs.
Accordingly, a graph convolution layer in GCN is divided into two propagation operations {\em Internal Propagation} and {\em Boarder Propagation} on the two decoupled local graphs respectively.

\begin{figure}[tbp]
	\centering
	\includegraphics[width=0.98\linewidth]{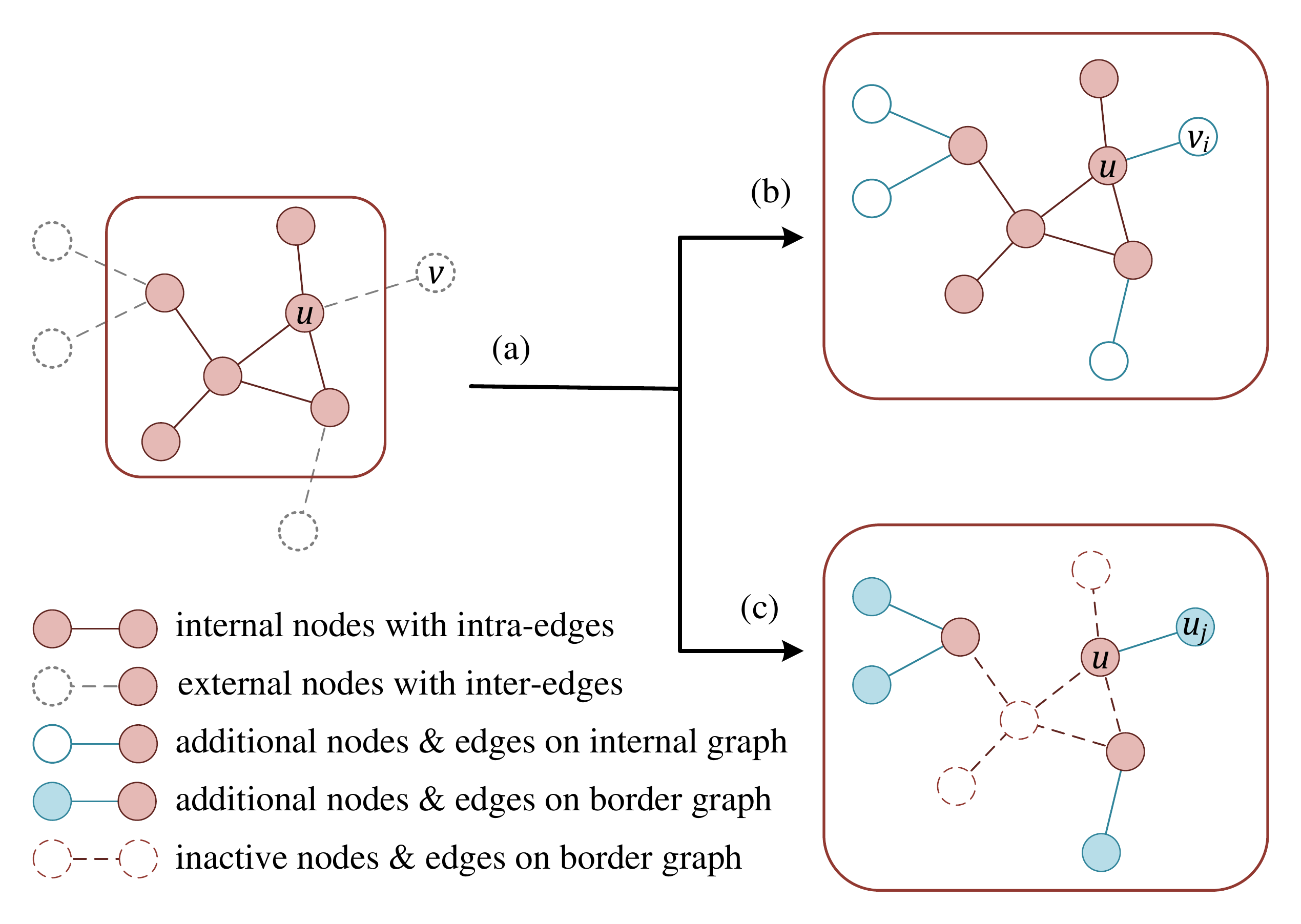}
	\vspace{-10pt}
	\caption{Graph decoupling of FedCog. (a) is the local graph of party $i$. The party does not have full knowledge of the inter-edges and some neighbor nodes. (b) and (c) show the decoupled graphs of (a). (b) is the \emph{internal graph}. The missing inter-edges and neighbor nodes are added to the local graph. (c) shows the \emph{border graph}, which also contains extra nodes and edges, but the edges between local internal nodes are removed. Thus the border graph is a bipartite graph of local nodes and extra nodes.}
	\label{fig:decoupling}
\end{figure}

\subsection{Graph Decoupling}\label{sec:decoupling}

The coupling of graphs causes data dependence among parties.
That is, a party cannot run GCN locally due to the lack of graph data held by other parties.
The first step of our method, FedCog, is proposed to eliminate the inter-party data dependence in GCN computation by decoupling the graphs.
Specifically, we decompose global graph $\CG$ into \emph{internal graphs} and \emph{border graphs} for the FL parties.

To build an \emph{internal graph} $\CG^I_i$,
party $i$ combines its local graph $\CG_i$ with the inter-edges which are connected to its internal nodes, and introduces additional nodes to complete the inter-edges.
For example, as shown in~\cref{fig:decoupling} from (a) to (b), $u\in V_i$ is party $i$'s internal node.
$v\in V_j\cap V^*_i$ is a node of party $j$'s local graph, and also an external border node of party $i$, i.e., $(u,v)$ is an inter-edge.
We introduce an additional node $v_i$ into the local graph as a substitute for node $v$, and connect $(u,v_i)$.
Seeing that party $i$ does not have knowledge about node $v$, the additional node $v_i$'s feature vector is initialized to zero.

The first step to build a \emph{border graph} $\CG^B_i$ is similar to the internal graph.
Additional nodes and edges are added to party $i$'s local graph $\CG_i$ according to the inter-edges.
Furthermore, party $i$ removes all intra-edges on its local graph $\CG_i$, and then the isolated internal nodes.
An instance is illustrated in~\cref{fig:decoupling} from (a) to (c).
Since an inter-edge $(u,v)$ connects party $i$'s internal node $u$ with another node $v$ on party $j$'s local graph $\CG_j$, we add a node $u_j$ as well as an edge $(u, u_j)$ to party $i$'s local graph.
After that, all intra-edges are removed.
The nodes with no inter-edges become isolated nodes on the border graph, and thus can be ignored.
Therefore, a border graph is a bipartite graph.
Each edge in $\CG^B_i$ connects an internal node and an additional node.
Both graphs $\CG^I_i$ and $\CG^B_i$ are locally held by party $i$ and then we will see that the graph decoupling facilitates FL over coupled graphs.

\subsection{Internal and Border Propagation}\label{sec:local_prop}

As the local graph $\CG_i$ is decoupled into two graphs, the graph convolution in GCN is accordingly divided into two sequential steps: \emph{Internal Propagation} and \emph{Border Propagation},
which are modified graph convolutions.

Internal Propagation is a graph convolution implemented on $\CG^I_i$:
\begin{equation}\label{eq:fedcog_prop1}
	\hat{\bh}_u\layer{l+\half}=\sum_{v\in\CN^I_u}\beta_{uv}\hat{\bh}_v\layer{l}\bW\layer{l},
\end{equation}
where $u$ is any node on internal graph $\CG^I_i$, including internal nodes in $V_i$ and the additional nodes introduced during graph decoupling.
The superscript $l+\half$ indicates the intermediate node representation after the first half of the two-step propagation.
Internal Propagation differs from the graph convolution in Eq.~\eqref{eq:gcn_prop} in two aspects:
1) a linear function is used as the activation function;
2) the weighted coefficient of neighbor representations is modified to $\beta_{uv}=1/\sqrt{1+d_v}$  rather than $c_{uv}$.
Note that this procedure is executed on graph $\CG^I_i$, and so the neighbors $\mathcal{N}^I_u$ and degrees $d_u$ are also defined on the decoupled graph $\CG^I_i$.

After Internal Propagation, each party sends the intermediate representations of internal nodes in $\CG^I_i$ (i.e. $h_{v_i}^{(l+\half)}$) to the party possessing node $v$.
This data transmission can be implemented on a decentralized fully-connected communication network as well as with the help of a centralized server.
Privacy analysis of this procedure is in Section~\ref{sec:privacy}.

Border Propagation is then executed on the bipartite local graph $\CG^B_i$.
\begin{equation}\label{eq:fedcog_prop2}
	\hat{\bh}_u\layer{l+1}=\sigma\left(\sum_{v\in\CN^B_u}\gamma_{uv}\hat{\bh}_v\layer{l+\half}\right), 
\end{equation}
where $\gamma_{uv}=1/\sqrt{1+d_u}$.
Here, $d_u$ is the degree of node $u$ in graph $\CG^I_i$, and $\CN^B_u$ is node $u$'s neighbors in graph $\CG^B_i$ with self-connections.

The following theorem demonstrates that our FedCog on coupled graphs
$\{\CG_i\}_{i=1}^{m}$ is equivalent to a graph convolution on global graph $\CG$.
\begin{theorem}
	\label{th:gcn_prop}
	Given a GCN on global graph $\CG=(V, E, X)$, i.e.,
	\[
	\bH\layer{l}=\operatorname*{GraphConv}(\bH\layer{l-1};\bW\layer{l-1}), \quad l=1,	\cdots, L,
	\]
	where $\bH\layer{0}=X$ and $\operatorname*{GraphConv}$ is an operation of applying Eq.~\eqref{eq:gcn_prop} to all nodes in $V$.
	Let $\hat{\bH}$ be the node representations obtained via Eq.~\eqref{eq:fedcog_prop1} and \eqref{eq:fedcog_prop2}.
	Then, we have 
	\[
	\hat{\bH}\layer{l}=\bH\layer{l}, \quad l=1,	\cdots, L.
	\]
\end{theorem}

\begin{proof}
	Let node $u\in V_i$ be an internal node of party $i$, and $u_j$ be an additional node on the border graph (see \cref{fig:decoupling} (c)).
	From~\cref{eq:fedcog_prop1}, we have
	\begin{equation}\label{eq:hatbhuj}
		\hat{\bh}_{u_j}\layer{l+\half}
		=\sum_{v\in\CN^I_{u_j}}\beta_{u_{j}v}\hat{\bh}_v\layer{l}\bW\layer{l}
	\end{equation}
	Note $B_u$ as the set of parties $j\neq i$ that hold node $u$ as an external border node, i.e., $B_u=\{l| l\in\{1, \ldots, m\}\setminus \{i\} \wedge u\in V_l^*\}$.
	Thus, 
	\begin{align}
		\label{eq:pf1_1}
		\hat{\bh}_u\layer{l+1}
		&=\sigma\left(\gamma_{uu}\hat{\bh}_{u}\layer{l+\half}+\sum_{j\in B_u}\gamma_{uu_j}\hat{\bh}_{u_j}\layer{l+\half}\right) \notag \\
		&=\sigma\left(\sum_{j\in B_u}\sum_{v\in\CN^I_{u_j}}c_{uv}\hat{\bh}_v\layer{l}\bW\layer{l}\right),
	\end{align}
	where $c_{uv}$ is defined in~\cref{eq:gcn_prop}.
	Remember the graph decoupling method, it is easy to find that the result in~\cref{eq:pf1_1} is the same as~\cref{eq:gcn_prop}.
\end{proof}

\subsection{Towards Efficient FedCog}\label{sec:efficient}

In the setting of federated learning, to train the parameters of neural networks, the communication cost for repeating FedCog in every training iteration is expensive.
Moreover, the backpropagation in gradient descent also requires the cooperation of parties, aggravating the communication cost.
We overcome these disadvantages by cleverly exploiting the properties of GCN variants, such as SGC~\cite{WuSZFYW19}, APPNP~\cite{KlicperaBG19}, GBP~\cite{ChenWDL00W20}, GPR-GNN~\cite{ChienP0M21}, etc.

As we mentioned in Section~\ref{sec:preliminaries}, SGC is a variant of GCN which removes the linear transforms and nonlinear activation functions between graph convolution layers.
Previous works suggest that SGC has a competitive performance over vanilla GCN~\cite{WuSZFYW19}.
A crucial feature of SGC when working with FedCog is the removal of trainable weights between graph convolution layers.
That is, the outputs of successive SGC layers are constant during the training of model weights.
Therefore, when SGC is adopted, we can compute the outputs of SGC layers only once, and then train model weights iteratively without repeating graph convolutions.
This allows FedCog to run the decoupling and propagation operations only once, instead of computing FedCog propagations in every training iteration.
After FedCog propagations, each party stores the embeddings of its nodes and learns the weights of neural networks via FedAvg~\cite{McMahanMRHA17}.
These processes are summarized in \cref{alg:fedcog_sgc}.

In \cref{alg:fedcog_sgc}, lines 2-3 are the graph decoupling procedures introduced in \cref{sec:decoupling}.
Each party applies the transform to its local graph $\CG_i$ and obtains an internal graph $\CG_i^I$ and a border graph $\CG_i^B$.
Lines 5-9 show the steps of internal and border propagations introduced in \cref{sec:local_prop}.
As we change the model from vanilla GCN to SGC, the propagation operation changes from \cref{eq:fedcog_prop1,eq:fedcog_prop2} into \cref{eq:fedcog_sgc_int,eq:fedcog_sgc_brd} accordingly
\begin{align}
	\label{eq:fedcog_sgc_int} \hat{\bh}_u\layer{l+\half}&=\sum_{v\in\CN^I_u}\beta_{uv}\bh_v\layer{l}, \\
	\label{eq:fedcog_sgc_brd} \hat{\bh}_u\layer{l+1}&=\sum_{v\in\CN^B_u}\gamma_{uv}\hat{\bh}_v\layer{l+\half}.
\end{align}
The correctness can be easily verified similar to~\cref{th:gcn_prop}.
When FedCog is deployed with SGC models, the propagation steps are executed only once.
Then in lines 11-17, we run the federated training steps, which will be explained in \cref{sec:fed_tr}.

\begin{algorithm}[t]
	\caption{FedCog-SGC}\label{alg:fedcog_sgc}
	\begin{algorithmic}[1]
		\Require{$m$ FL parties with their local subgraphs $\{\CG_i\}_{i=1}^m$, $L$-layer SGC model $f(\CG; \theta)$, $\theta=\{\bW\layer{0},\cdots,\bW\layer{L-1}\}$}
		\LineComment{Graph Decoupling}
		\For{$i=1$ to $m$}
		\Comment{Execute on party $i$}
		\State $\CG^I_i, \CG^B_i\leftarrow\operatorname*{GraphDecoupling}(\CG_i)$
		\EndFor
		\LineComment{Internal and Border Propagations}
		\For{$l=0$ to $L-1$}
		\For{$i=1$ to $m$}
		\Comment{Execute on party $i$}
		\State $\hat{\bh}_u\layer{l+\half}\leftarrow\sum_{v\in\CN^I_u}\beta_{uv}\bh_v\layer{l}$
		\State Receive $\left\{\hat{\bh}_{u_v}\layer{l+\half}: v\in V_j\right\}$ from party $j$
		\State $\hat{\bh}_u\layer{l+1}\leftarrow\sum_{v\in\CN^B_u}\gamma_{uv}\hat{\bh}_v\layer{l+\half}$
		\EndFor
		\EndFor
		\LineComment{Federated Training}
		\For{$t=1$ to $k$}
		\State Sample a set of parties $\CC_t$
		\For{$i\in\CC_t$}
		\Comment{Execute on party $i$}
		\State $\theta_i\leftarrow\theta$
		\State 
		$\boldsymbol{g}_i\leftarrow\operatorname*{GradientDescent}(\theta_i,\{\hat{\bh}\layer{L}_u\}_{u\in V_i})$
		\EndFor
		\State $\bar{\boldsymbol{g}}\leftarrow\sum_{i\in\CC_t}p_i\layer{t}\boldsymbol{g}_i$
		\Comment{Execute on FL Server}
		\State $\theta\leftarrow\theta-\eta\bar{\boldsymbol{g}}$
		\EndFor
	\end{algorithmic}
\end{algorithm}

In addition to SGC, some more GNN models also output constant graph convolution results during model training and work efficiently with FedCog.
APPNP~\cite{KlicperaBG19} uses the predict-then-propagate framework to build
neural networks, where its propagation is achieved by approximating Personalized
PageRanks.
The major procedures of FedCog-APPNP are similar to SGC.
The only difference occurs at Border Propagation.
It modifies~\cref{eq:fedcog_sgc_brd} to 
\begin{equation}
	\hat{\bh}_u\layer{l+1}=(1-\alpha)\sum_{v\in\CN^B_u}\gamma_{uv}\hat{\bh}_v\layer{l+\half}+\alpha\hat{\bh}_u\layer{0},
\end{equation}
where $\alpha\in (0,1]$ is the restart probability of Personalized PageRank.

GBP~\cite{ChenWDL00W20} extends the feature propagating method in SGC to
Generalized PageRank (GPR)~\cite{0005CM19}.
The propagation is computed through $\bH^{(l+1)} =
\tilde{\bD}^{r-1}\tilde{\bA}\tilde{\bD}^{-r}\bH^{(l)}, r\in[0,1]$.
Therefore, to apply our method to GPR,  we modify the coefficients in~\cref{eq:fedcog_sgc_int,eq:fedcog_sgc_brd} to
\begin{align}
	\beta_{uv}&=\frac{1}{(1+d_v)^{1-r}}, \\
	\gamma_{uv}&=\frac{1}{(1+d_u)^{r}}.
\end{align}

GPR-GNN~\cite{ChienP0M21} is also a GNN model motivated by GPR.
From the computation perspective, GPR-GNN can be viewed as applying a weighted
sum to the hidden features of SGC.
Therefore, our method is also feasible for GPR-GNN.

We take the SGC model as an example to analyze the computation and communication complexities of our method FedCog.
Suppose there are $m$ parties collaborating to train the FL model.
Each party keeps the graph data with $n$ nodes, $e_I$ intra-edges and $e_E$
inter-edges.
The node features in the graphs are embedded in the $F$-dim space.
When we use a GNN model with $L$ graph convolutional layers and apply $k$
iterations of federated training, the computation and communication costs are summarized in \cref{tab:eff_theory}.

From the aspect of computation cost, FedCog utilizes the information of inter-edges, which is not involved in original SGC-based Federated Learning. The computation cost of FedCog is larger than SGC during propagation,
but FedCog keeps the same efficiency as SGC in training steps, which is faster than GCN.
As for the communication cost, FedCog makes a coupled feature propagation and leads to a higher cost than SGC.
The communication cost of FedCog is in proportion to the size of graph $n$, so it will be expensive to apply FedCog to large graphs.
Despite this, our later experiments show that such costs are not extremely expensive, which is acceptable in practice.

\begin{table}[t]
	\caption{Costs of graph FL methods}
	\label{tab:eff_theory}
	\vspace{-2ex}
	\begin{tabular}{ccc}
		\hline
		Method  & Computation           & Communication \\ \hline
		GCN & $O(kmLe_IF+kmLnF^2)$     & $O(kmLF^2)$      \\
		SGC & $O(mLe_IF+kmnF^2)$       & $O(kmF^2)$       \\
		FedCog-SGC & $O(mL(e_I+e_E)F+kmnF^2)$ & $O(mnF+kmF^2)$   \\ \hline
	\end{tabular}
	\vspace{-2ex}
\end{table}

\subsection{Federated Training}\label{sec:fed_tr}

FedCog is designed as a substitute propagation method for GNN models and does not affect the model parameters' local and global update steps during federated training.
Therefore, FedCog can be implemented as a plug-in algorithm for various FL methods.
For instance, GNN parameters can be trained with FedAvg~\cite{McMahanMRHA17}.
The procedure is shown in lines 11-17 in Algorithm~\ref{alg:fedcog_sgc}.
We can also apply FedCog to other FL training frameworks, such as FedDyn\cite{AcarZNMWS21} and FedOpt\cite{ReddiCZGRKKM21} to improve the model convergence, especially when statistical heterogeneity occurs in parties' local graph data.

\subsection{Privacy Preservation}\label{sec:privacy}

The coupled graph problems often arise among company or organization parties.
The parties are generally honest and trusted.
However, the risk of malicious adversarial parties should never be neglected.
The propagation and federated training may cause privacy leakage to the adversarial server and parties.
We analyze the privacy security of our FedCog in this section. 

\bullethdr{Privacy Security of Propagation.}
During the propagation procedures, FedCog requires the FL parties to exchange some intermediate computation results.
In Section~\ref{sec:local_prop}, we mentioned two types of information exchange: a decentralized peer-to-peer exchange, and a centralized exchange through a server.
Next, we will first introduce the weakness of FedCog with the decentralized propagation, and propose a method, named \emph{LNNC}, to address this problem.
Then, we will show that LNNC also solves the privacy problem in the centralized manner.

In practical coupled graph scenarios, a node's feature vector is usually sensitive information.
For example, a feature vector may represent attributes of a user on an online social network.
Therefore, FedCog must ensure that a malicious party or server cannot obtain any node feature vectors during the propagation.
That is, our goal is to guarantee that any entry of party $i$'s node features $\bX_i$ will never be revealed by the server and other parties during FedCog.

In decentralized propagation, we consider a set of parties $\CC_A$ as malicious adversaries.
In our threat model, we assume that all adversarial parties are semi-honest and they may collude with each other.
An adversarial party $i$ has prior knowledge $\{\CG_i, \CH_j\}$, where $\CG_i=(V_i, V^*_i, E_i, \bX_i)$ is party $i$'s local graph data, and $\CH_j=\{\hat{\bh}_{u_j}\layer{l+\half}|u\in V_i\cap V_j^*, l=0,1,\ldots\}$ is the intermediate results received from a victim party $j$.

Adversarial party $i$'s goal is to infer any node feature vector $\bx_v$, $v\in V_j$ on the victim party $j$.
Thus, the privacy scheme must avoid the leakage of party $j$'s any node feature.
We propose to protect the confidentiality of individual node features based on the following theorem.

\begin{theorem}\label{th:lnnc}
	Given a victim party $j$'s internal node $v\in V_j$.
	The necessary condition of collusive adversaries $\CC_A$ obtaining the value of $\bx_{v}$ is
	$\exists u\in V_j$, s.t. $\RCN_u\subseteq V_{A}$,
	where $V_{A}=\bigcup_{i\in \CC_A}V_i$ is the union of all adversaries' internal nodes, and $\RCN_{u}=\CN_u\setminus\{u\}$ is the set of node $u$'s neighbors on graph $G$ without self-connections.
\end{theorem}
\begin{proof}
	
	Consider the worst situation of~\cref{th:lnnc}, i.e., all parties except a victim party are adversaries and collusive.
	Then, we regard the union of all adversarial parties $\CC_A$ as one party and simplify the case to a two-party scenario: Party A is the malicious party and  Party B is the victim party.
	Each Party $i$ holding a local graph $\CG_i=(V_i, V_i^*, E_i, \bX_i)$, $i=A,B$.
	
	\textbf{(1) In SGC layer 1:}
	For any external node $u\in V_B^*$ of the victim, we have
	\begin{equation}\label{eq:pf2_layer1}
		\hat{\bh}_{u_B}\layer{\half}
		=\sum_{v\in\RCN_{u_B}^I}\beta_{u_{B}v}\bx_{v}.
	\end{equation}
	The adversary has no knowledge of the victim's intra-edges, therefore variable $\beta_{u_B v}$ is unknown to the adversary.
	Let $F$ be the number of node feature attributes.
	There are $F$ equations in~\cref{eq:pf2_layer1}, which might be utilized by the adversary to infer sensitive information about the victim, but the number of unknown variables $\beta_{u_B v}, \bx_{v}, \forall v\in\RCN_{u_B}^I$ is $|\RCN_{u_B}^I|(F+1)\geq F$.
	Furthermore, when selecting any $r$ equations from the $F$ equations in~\cref{eq:pf2_layer1}, there are $|\RCN_{u_B}^I|(r+1)>r$ independent variables.
	Therefore, the adversary fails to infer any entries in $\bx_{v}$ via $\hat{\bh}_{u_B}\layer{\half}$.
	
	\textbf{(2) In SGC layer 2:}
	For any external node $u\in V_B^*$ of the victim, we have
	\begin{align}\label{eq:pf2_layer2_3}
		\hat{\bh}_{u_B}\layer{1+\half}
		&=\sum_{v\in\RCN_{u_B}^I}\beta_{u_{B}v}\hat{\bh}_{v}\layer{1} \notag \\
		&=\sum_{v\in\RCN_{u_B}^I}\beta_{u_{B}v}\gamma_{vv}\sum_{w\in\RCN_{v}\cap V_B}\beta_{vw}\bx_{w} \notag \\
		&+\sum_{v\in\RCN_{u_B}^I}\beta_{u_{B}v}\gamma_{vv}\beta_{vv}\bx_{v}
		+\sum_{v\in\RCN_{u_B}^I}\beta_{u_{B}v}\gamma_{vv_A}\hat{\bh}_{v_A}\layer{\half}.
	\end{align}
	Combining~\cref{eq:pf2_layer1,eq:pf2_layer2_3}, the adversary has $2F$ equations, and the total number of independent unknown variables is
	\begin{equation}\label{eq:pf2_total}
		T=|\RCN_{u_B}^I|(F+2)+\sum_{v\in\RCN_{u_B}^I}|\RCN_{v}\cap V_B|(F+1).
	\end{equation}
	If the necessary condition in~\cref{th:lnnc} does not hold, i.e., $|\RCN_{v}\cap V_B| \geq 1, \forall v\in V_B$, then we have
	\begin{align}
		T
		&=|\RCN_{u_2}^I|(F+2)+\sum_{v\in\RCN_{u_2}^I}|\RCN_{v}\cap V_2|(F+1) \notag \\
		&\geq |\RCN_{u_2}^I|(F+2)+\sum_{v\in\RCN_{u_2}^I}(F+1)  > 2F.
	\end{align}
	If we select any $r$ equations from~\cref{eq:pf2_layer1} and $s$ equations from~\cref{eq:pf2_layer2_3}, the number of independent variables in these selected equations is $|\RCN_{u_B}^I|(\max(r,s)+2)+\sum_{v\in\RCN_{u_B}^I}|\RCN_{v}\cap V_B|(s+1)>r+s$.
	Therefore, the adversary still fails to infer any node features $\bx_v$, $v\in V_B$ from $\hat{\bh}_{u_B}\layer{\half}$ and $\hat{\bh}_{u_B}\layer{1+\half}$.
	
	\textbf{(3) In SGC layer $L\geq3$:} The adversary has $LF$ equations with respect to $\{\bx_{v}\}_{v\in V_B}$.
	However, similar to the steps in (2), we have at least $LF+2L-1$ independent variables to be solved when the condition in~\cref{th:lnnc} does not hold.
	Therefore, the node features possessed by the victim Party $B$ cannot be inferred by the adversary Party $A$.
\end{proof}

\cref{th:lnnc} reveals that the intermediate results $\CH_j$ may be insecure in some scenarios.
In addition, it also inspires us to guarantee privacy security by breaking the necessary condition.
\cref{th:lnnc} indicates that the adversary can obtain party $j$'s private information only when there are nodes in $V_j$ which have no internal neighbors, i.e. no intra-edges, on the local graph $\CG_j$.
Thus, we propose {\em Local Nearest Neighbor Connection} (LNNC) to protect privacy security by adding intra-edges to the nodes without internal neighbors.

The procedure of LNNC is as follows.
Before party $j$ starts FedCog propagations, it retrieves its internal nodes in $V_j$ which satisfy $\RCN_u\subseteq V_{j}^*$. Note that $V_{j}^* \subseteq V_{A}$.
For each retrieved node $u$, party $j$ selects another internal node $u_N\in V_j\setminus\{u\}$ and adds an edge $(u, u_N)$ to graph $\CG_j$.
We notice that propagation is a weighted sum operation.
Intuitively, adding an additional edge which is connected to a node with similar features will have a slight influence on the propagation results.
Therefore, we select the node $u_N$ as the nearest node in the feature space,
i.e., $u_N = \arg\min_{v\in V_j\setminus\{u\}}D(\bx_u, \bx_v)$, where
$D(\cdot,\cdot)$ is a distance metric.
In this work, we use the angular distance, i.e., $D(\bx,\by) =
\frac{1}{\pi}\cos^{-1}\left(\frac{\bx\transpose \by}{\|\bx\|\|\by\|}\right)$.
Using LNNC, we guarantee that each internal node of party $j$ has at least one intra-edge.
Therefore, we broke a necessary condition of privacy leakage.
After using LNNC, when we start the propagating and training procedures of FedCog, the adversarial parties will not get the sensitive data.

Besides the decentralized transmission, we also consider a centralized form to exchange intermediate data.
A server receives $\hat{\bh}_{v_i}\layer{l+\half}$ and forwards it to party $i$.
In general, the server cannot access parties' local data, so its prior knowledge is generally less than the adversarial parties.
In the worst situation, the server may collude with adversarial parties, but it still cannot get more prior knowledge than the adversarial parties.
Thus, privacy is preserved from the server as long as from the parties.
In centralized transmission, the adversarial parties have the same prior knowledge as the decentralized scenario, so the LNNC method is still effective to resolve privacy risks.
Therefore, LNNC also guarantees privacy security in centralized transmission.

Notice that the correctness of~\cref{th:lnnc} is established on the secrecy of $\beta_{uv}$ in~\cref{eq:fedcog_prop1}, which depends on the GNN models.
Thus, LNNC cannot ensure data privacy for some GNN models, e.g., GraphSage with a mean ($\beta_{uv}$ is constant) or pooling ($\beta_{uv}\in\{0,1\}$) aggregators~\cite{HamiltonYL17}, because the attacker may obtain $\beta_{uv}$ without prior knowledge of the victim's local graph structure.
This is a limitation of FedCog and LNNC, while LNNC can still provide a privacy guarantee for a large number of widely used GNN models, such as GCN~\cite{KipfW17}, SGC~\cite{WuSZFYW19}, APPNP~\cite{KlicperaBG19}, GBP~\cite{ChenWDL00W20}, GPR-GNN~\cite{ChienP0M21}, etc.

\bullethdr{Privacy Security of Federated Training.}
In the training steps, the FL server collects gradients $\boldsymbol{g}_i^{(t)}$ from parties, which may also increase the risk of privacy leakage.
To address this, homomorphic encryption such as the Paillier cryptosystem~\cite{Paillier1999} can be applied to achieve secure computations on the untrusted server, which prevents the untrusted server from accessing private information.
In addition, previous works, such as~\cite{BonawitzIKMMPRS17, abs-2009-11248} providing secure aggregation methods can also be used to preserve privacy in the training steps of our FedCog.

\section{Evaluation} \label{sec:results}

In this section, we evaluate our method FedCog on node classification and link prediction tasks.
We show empirical results of model performance and efficiency.
In our experiment, all FL servers and parties were simulated by processes on a physical device with Intel Xeon Silver 4316 CPU and NVIDIA Tesla V100S GPU.

\begin{table}[tbp]
	\centering
	\caption{Summary of real-world datasets}
	\label{tab:real_dataset}
	\vspace{-5pt}
	\begin{tabular}{ccccc}
		\hline
		dataset      & \#nodes & \#edges & \#classes & \#attributes \\ \hline
		CORA~\cite{LuG03}         & 2,708    & 5,429    & 7          & 1,433         \\
		CiteSeer~\cite{SenNBGGE08}     & 3,327    & 4,732    & 6          & 3,703         \\
		PubMed~\cite{SenNBGGE08}       & 19,717   & 44,338   & 3          & 500           \\
		DBLP~\cite{PanWZZW16}       & 17,716   & 52,867   & 4          & 1,639           \\
		Coauthor-CS~\cite{shchur2018pitfalls}  & 18,333   & 81,894   & 15         & 6,805           \\
		Coauthor-Phy~\cite{shchur2018pitfalls} & 34,493   & 247,962  & 5          & 8,415           \\
		ACM~\cite{han2019} & 3,025   & 13,128  & 3          & 1,870           \\
		BlogCatalog~\cite{LiHTL15} & 5,196   & 171,743  & 6          & 8,189           \\
		Flickr~\cite{LiHTL15}       & 7,575    & 239,738  & 9          & 12,047        \\
		Amazon-CS~\cite{McAuleyTSH15}  & 13,752   & 287,209   & 10         & 767         \\
		Amazon-Photo~\cite{McAuleyTSH15} & 7,650   & 143,663  & 8          & 745         \\
		UAI~\cite{wang2018unified} & 3,067   & 28,311  & 19          & 4,973         \\ \hline
	\end{tabular}
	\vspace{-10pt}
\end{table}

\subsection{Experimental Setup}

\noindent$\bullet$ \textbf{Datasets.} 
Table~\ref{tab:real_dataset} summarizes the 12 real-world datasets from 5 domains used for evaluating the performance of our method.
CORA~\cite{LuG03}, CiteSeer, PubMed~\cite{SenNBGGE08}, and DBLP~\cite{PanWZZW16} are citation networks.
Coauthor-CS, Coauthor-Phy~\cite{shchur2018pitfalls}, and ACM~\cite{han2019} represent co-authorship relations between authors and papers.
BlogCatalog and Flickr~\cite{LiHTL15} are social networks.
Amazon-CS and Amazon-Photo~\cite{McAuleyTSH15} are graphs of goods that are usually bought together.
UAI~\cite{wang2018unified} is a graph of Internet document network including the reference relations.
To simulate the situation of coupled graphs, we partitioned the datasets into subgraphs as the preprocess.
Each graph was partitioned into $\{2, 5, 10, 20, 50, 100\}$ parts, and each part was distributed to one party.

\begin{figure*}[tbp]
	\centering
	\includegraphics[width=0.85\linewidth]{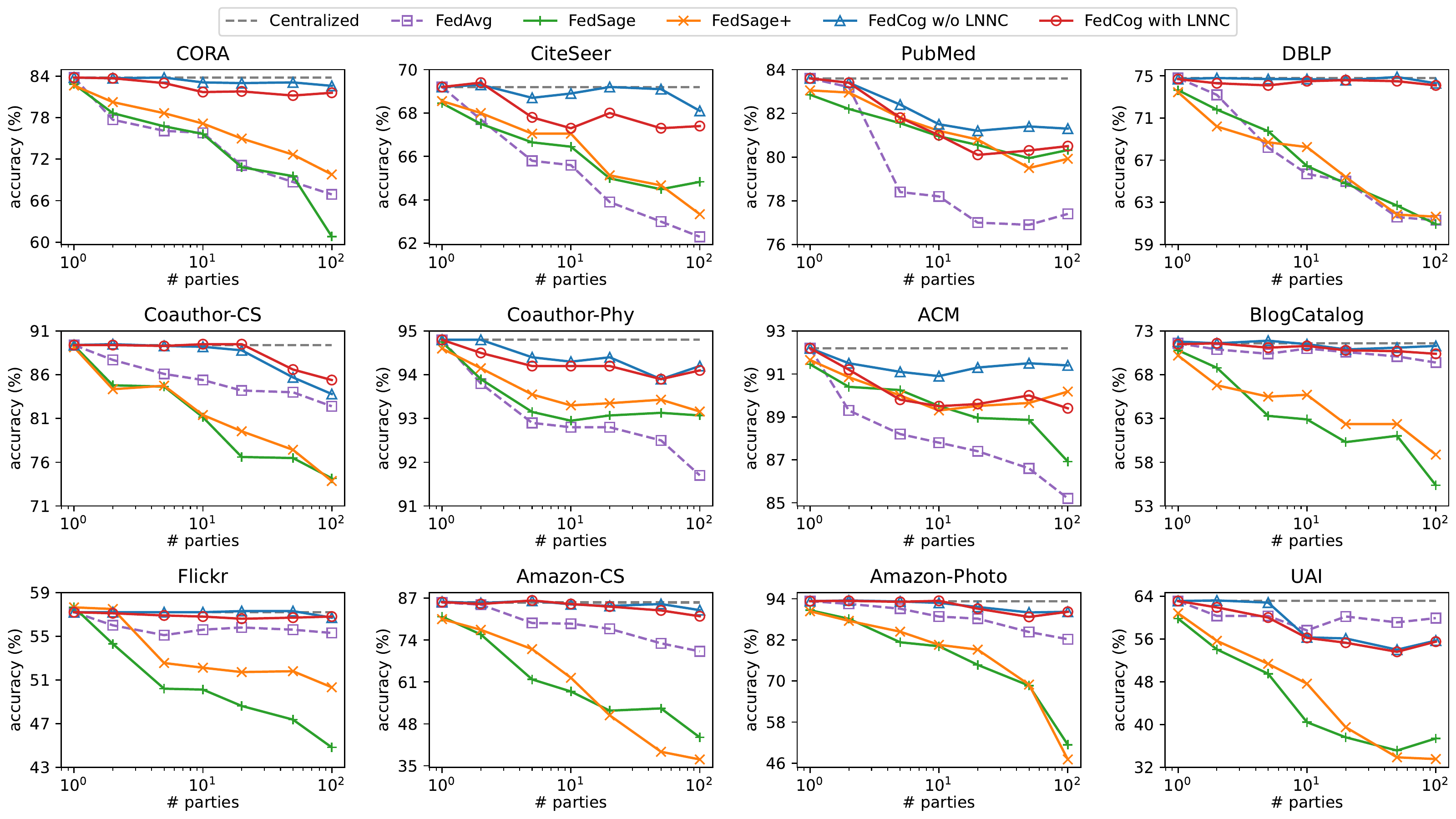}
	\caption{Node classification accuracy when the global graphs are partitioned by K-Means.}
	\label{fig:nc_kmeans}
\end{figure*}

We adopted two partitioning methods to evaluate FedCog's performances in different local data distributions comprehensively: \emph{topological clustering} and \emph{feature clustering}.
We use METIS~\cite{KarypisK98} algorithm to partition graphs according to their topological structures.
Feature clustering is to cluster the nodes into groups by applying K-Means to node features.
Both partitioning methods generate subgraphs with no nodes in common.
An important difference between the two partitioning methods is the number of intra-edges in the local graphs $\CG_i$.
The topological clustering algorithm will try to keep high edge density in the partitioned subgraphs.
Thus the local graphs will have more intra-edges than local graphs partitioned by other methods.
However, the feature clustering method will ignore the topological structure during partitioning.
Feature clustering may lead to sparse local subgraphs.
For example, when we used METIS to cut CORA and CiteSeer graphs into 100 parts, original graphs' 54.62\% and 79.62\% edges became subgraphs' intra-edges respectively.
Our LNNC method only added 0.21\% and 0.02\% edges compared with the original graphs.
In the case of K-Means for 100-subgraph partitioning, only 27.32\% and 36.23\% of edges were intra-edges in CORA and CiteSeer datasets respectively.
Our LNNC method added 9.87\% and 9.26\% edges to satisfy privacy conditions.

The difference in local data distributions is also reflected in the unbalanced labels of local data.
In general, the neighbor nodes on a graph likely have the same label, and so do the nodes with similar features.
Therefore, both partitioning methods tend to cluster the nodes with the same labels into one party.
This will cause unbalanced label distributions among parties.
We use average earth mover distance (EMD)~\cite{HsuQ020} of label distributions to measure the unbalance.
EMD is in the range of $[0, 2]$, where a larger EMD means more unbalanced.
If each dataset is partitioned into 100 parts, the average EMD over 12 datasets is 1.29 for METIS and 1.46 for K-Means, which shows a degree of unbalance.
In particular, the average EMD of CORA, Flickr and UAI datasets under K-Means partitioning is 1.69, 1.70 and 1.71, respectively.
It indicates that these datasets have highly unbalanced local data.

\noindent$\bullet$ \textbf{Baselines.} To show the effectiveness of FedCog, we compare the performance of  FedCog with the regular FedAvg, FedSage, and FedSage+\cite{ZhangYLSY21}.
The regular FedAvg method is a simple combination of SGC~\cite{WuSZFYW19} and Federated Learning method based on~\cite{McMahanMRHA17}.
In our experiments, FedAvg and FedSage regard each party's local subgraph as an individual graph, ignoring the existence of inter-edges.
FedSage+ trains a neighbor generator for each party and complements local graphs.
FedSage+ utilizes both local and foreign data, so it is an available solution for coupled graph problems.
As for our method FedCog, we adopt SGC as the GNN model for node classification and link prediction tasks.
SGC models in the experiments contain 2 simplified graph convolutional layers.
In addition, we use a centralized SGC model as a reference method.
Centralized SGC collects local data from all parties and trains the model on a server without regard to data privacy, which is an ideal but impractical setting for learning tasks.
For all methods, we tune the learning rate parameter in the range of $[0.001, 0.1]$ during training.

We also conduct experiments to show FedCog's applicability to different federated training algorithms.
FedAvg~\cite{McMahanMRHA17}, FedDyn~\cite{AcarZNMWS21}, FedAdagrad and FedAdam~\cite{ReddiCZGRKKM21} are selected as baselines.
FedCog is integrated into the four FL training algorithms and compared with the FedCog-free versions.
In this experiment, the settings of SGC models are the same as the above.
During the federated training of all baselines and our method, we iteratively apply 1 epoch of local model training and one round of global model update.
For all four methods, learning rates are tuned in $[0.001, 0.1]$.
FedDyn's parameter $\alpha$ is selected in $[0.001, 0.1]$.
For FedAdagrad and FedAdam, we set parameter $\tau\in[10^{-5}, 10^{-1}]$.
As~\cite{ReddiCZGRKKM21} suggested, we set $\beta_1=0.9$ and $\beta_2=0.99$ for FedAdam.

\noindent$\bullet$ \textbf{Metrics.} In node classification tasks, we use micro-average precision of predictions to evaluate the accuracy performance of the methods.
Larger accuracy values are better results.
As for link prediction tasks, we use the area under the ROC curve (AUC) to show the results as a measurement metric.
ROC is the TPR-FPR curve of a binary classifier, where $\text{TPR}=\frac{\text{TP}}{\text{TP}+\text{FN}}$, $\text{FPR}=\frac{\text{FP}}{\text{FP}+\text{TN}}$.
Larger AUC scores indicate the model has better performance.

\subsection{Results of Node Classification}\label{sec:res_nc}

\begin{figure*}[tbp]
	\centering
	\includegraphics[width=0.85\linewidth]{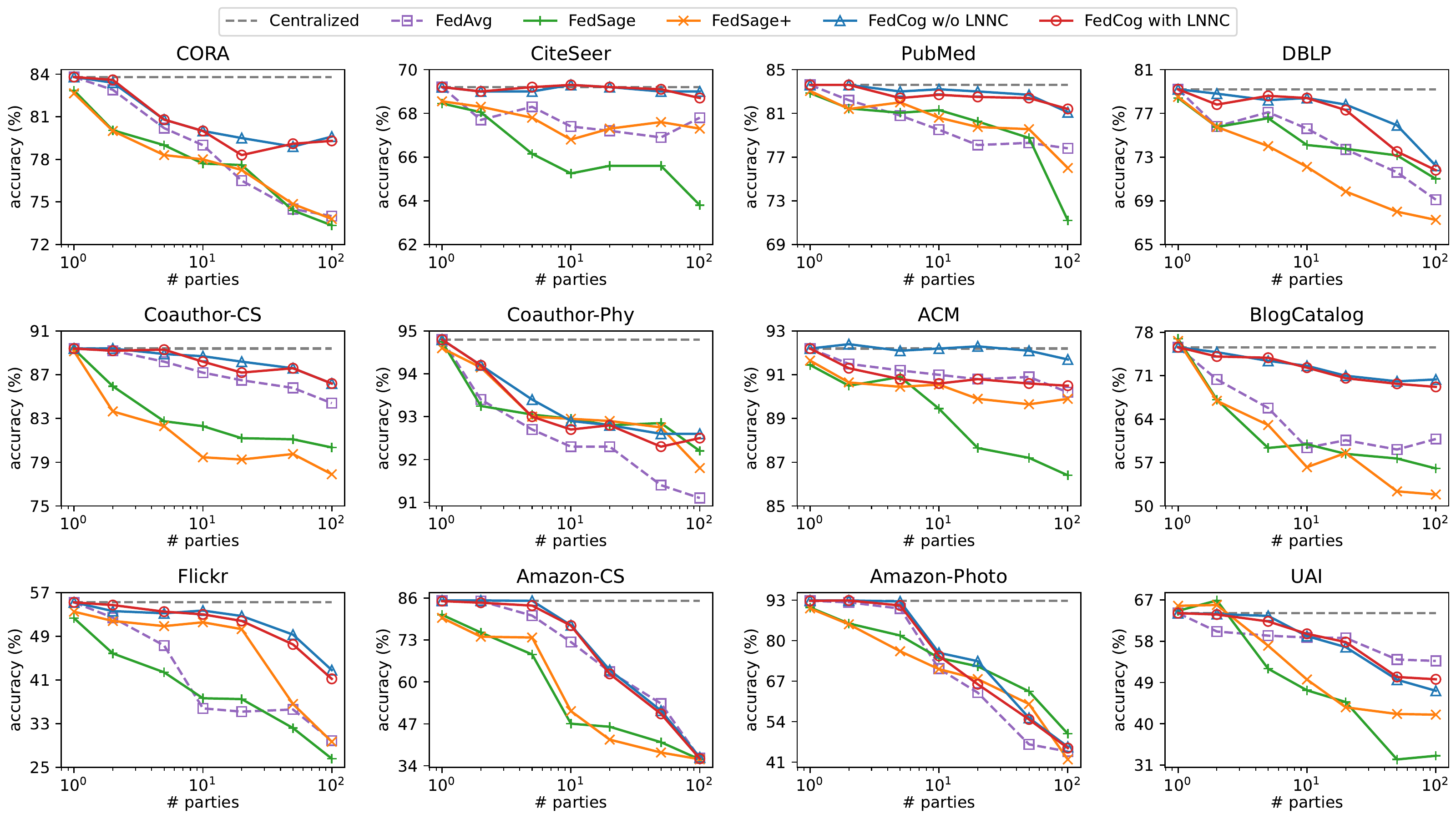}
	\vspace{-10pt}
	\caption{Node classification accuracy when the global graphs are partitioned by METIS.}
\label{fig:nc_metis}
\end{figure*}

We examine the node classification performance in this experiment.
For each dataset, we sample 30 nodes per class as the training set, and then uniformly sample 1,000 nodes other than training samples as the test set.
METIS and K-Means are used to partition the graphs into at most 100 parts as local graph data.
We achieve federated node classification tasks by baselines FedAvg, FedSage, FedSage+, and our method FedCog.
In the experiments for FedSage and FedSage+, we follow the settings explained in~\cite{ZhangYLSY21}.
For the FedAvg method, the GNN model contains 2 layers of Simplified Graph Convolution followed by a fully-connected layer.
As for FedCog, model settings are similar to that in FedAvg.
The difference is that we repeat the propagations of FedCog-SGC 2 times to simulate a 2-layer SGC on the global graph.

\cref{fig:nc_kmeans} illustrates node classification accuracy when the graphs are partitioned by K-Means node feature clustering.
\emph{FedAvg}, \emph{FedSage} and \emph{FedSage+} show the baseline results respectively, and \emph{FedCog with LNNC} indicates the results of our method.
To show the influence of adding intra-edges during LNNC, we also illustrate the results of \emph{FedCog w/o LNNC} as a reference.
Besides, we also conduct experiments on the entire graphs without partitioning, which is noted as \emph{Centralized} in the results.

From the results of baseline FedAvg, we observe that on most datasets, such as CORA, DBLP, and Amazon-CS, FedAvg has a significant accuracy decrease as the number of parties increases.
As a comparison, our FedCog method has better performance than FedAvg, and even maintains almost the same accuracy as no partitioning on some datasets, such as CORA and DBLP.
When there are 100 parties, our FedCog's accuracy is 14.7\%, 12.8\%, and 10.9\% higher than FedAvg on CORA, DBLP, and Amazon-CS respectively.
Although on PubMed and ACM datasets, FedCog's performance also declines along with party number growth, our method still outperforms FedAvg.
Compared with FedSage and FedSage+ baselines, FedCog's accuracy is similar to FedSage+ on PubMed and ACM datasets.
FedCog outperforms FedSage+ on the other 10 datasets.

When we compare FedCog with its LNNC-free version, we find that LNNC slightly weakens classification accuracy.
In our experiments, the maximum accuracy decrease caused by LNNC occurs on the 100-party ACM dataset.
LNNC causes an accuracy decrease by 2.0\%, which is a tolerable cost to guarantee privacy security. 

\begin{figure*}[tbp]
	\centering
	\includegraphics[width=0.85\linewidth]{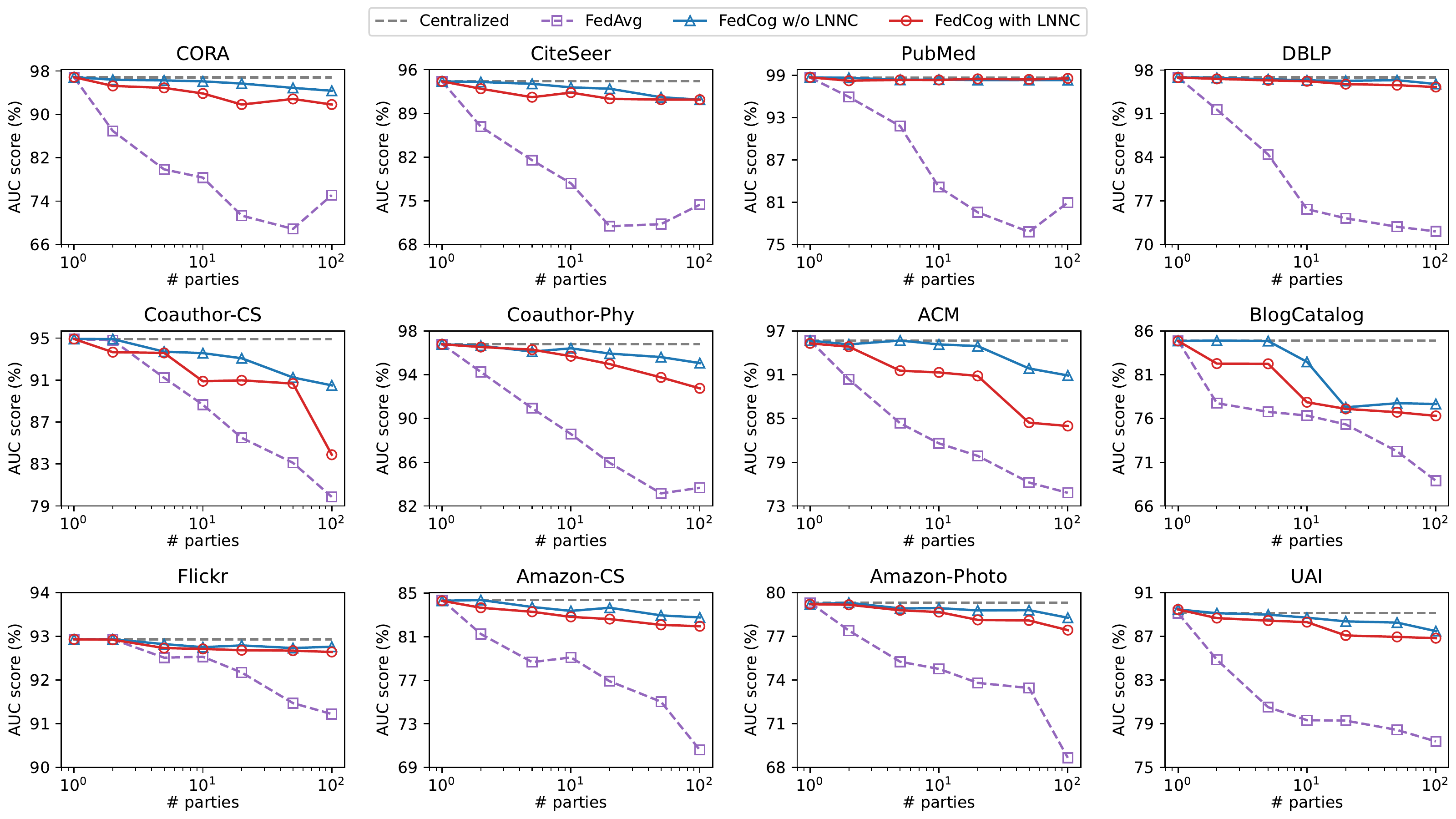}
	\vspace{-10pt}
	\caption{AUC scores of link prediction tasks when the global graphs are partitioned by K-Means.}
	\label{fig:lp_kmeans}
\end{figure*}

\cref{fig:nc_metis} shows node classification accuracy when the graphs are partitioned by METIS.
In these experiments, our FedCog also outperforms FedAvg when the graphs are partitioned into various parties.
On CORA, BlogCatalog, and Flickr datasets, the results of FedCog are 5.3\%, 8.4\%, and 11.3\% better than FedAvg at 100 parties.
FedAvg performs better than FedCog on the UAI dataset, but such results do not appear on the rest 11 datasets.
When we compare FedCog with FedSage and FedSage+, we observe that our method has comparable results on some datasets, such as Coauthor-Phy, ACM, and Amazon-Photo.
There are also some datasets where our method FedCog performs better, such as CORA and PubMed.
LNNC has no significant influence on the classification results over most datasets except ACM.
LNNC causes a 1.2\% accuracy decrease on the ACM dataset, and consequently, FedCog does not outperform FedAvg on this dataset.
However, on all of the other datasets, LNNC does not influence the performance significantly.
Thus LNNC is an acceptable solution for privacy preservation.

In addition, we would like to point out that graph partitioning methods have a significant impact on classification accuracy even if the original dataset is the same.
For example, if we change the partitioning method from METIS to K-Means for 100 parties, the accuracy changes of FedAvg are -7.1\% on Cora and -5.5\% on CiteSeer.
These results show that when graphs are cut by different partitioning methods (or say when the distributions of local subgraphs are different), FedAvg will be significantly affected.
However, the corresponding results of FedCog are +2.3\% on Cora and -1.3\% on CiteSeer respectively.
The change is much smaller than FedAvg, which suggests that our FedCog method would perform better in the highly non-IID federated tasks of real-world scenarios.

\subsection{Results of Link Prediction}\label{sec:res_lp}

We also conduct experiments on link prediction tasks.
We use K-Means to partition the graphs into at most 100 parts as local graph data.
For each dataset, we sample two sets of edges as the training and test sets successively.
In both sets, the proportions of positive and negative samples are 1:1.
In our experiment, the neural network model is an SGC model followed by a dot predictor.
The 2-layer SGC outputs a 100-dim embedding for each node.
Then the dot predictor takes a pair of node embeddings as its input and outputs a predictive score by computing the inner product of the two embedding vectors.
We compute AUC scores based on the predictive scores to evaluate each method's performance.

The results of link prediction are shown in~\cref{fig:lp_kmeans}.
The finding is similar to node classification: the performance of FedAvg decreases along with the increase of parties, while the performance of our FedCog changes slightly.
The results of FedCog are better than the baselines when the graphs are cut into a large number of parties.
When the number of parties is 100, FedCog outperforms FedAvg by 0.168, 0.176, and 0.232 on CiteSeer, PubMed, and DBLP datasets respectively.
These results show that our FedCog method is better than the FedAvg baseline in link prediction tasks.
LNNC scheme causes performance degradation on some datasets, such as Coauthor-CS and ACM.
The AUC scores decrease by 0.066 and 0.069 on Coauthor-CS and ACM datasets respectively.
However, FedCog's results are still better than the baseline methods.
Therefore, FedCog is an effective approach to improve link prediction results on coupled graphs.

\subsection{Results of Efficiency}\label{sec:res_eff}

We test the computation and communication efficiency of FedCog on 4 large-scale datasets from different domains: DBLP, Coauthor-Phy, Flickr, and Amazon-CS.
The settings of GNN models and FL training algorithms are the same as the node classification experiments in~\cref{sec:res_nc}.
FedCog only changes the propagation procedure of SGC, so we simply compare the computation and communication costs during the propagation phase (parameter updates are excluded) for FedCog and all baselines.

\begin{table*}[tbp]
	\centering
	\caption{Run-times on 100 FL parties}
	\label{tab:eff_compute}
	\vspace{-10pt}
	\begin{tabular}{c|cccc|cc|c}
		\hline
		&
		\begin{tabular}[c]{@{}c@{}}FedCog\\ (ms)\end{tabular} &
		\begin{tabular}[c]{@{}c@{}}FedAvg\\ (ms)\end{tabular} &
		\begin{tabular}[c]{@{}c@{}}Centralized\\ (ms)\end{tabular} &
		\begin{tabular}[c]{@{}c@{}}Param. Update\\ (ms/round)\end{tabular} &
		$\frac{\rm FedCog}{\rm Centralized}$ &
		$\frac{\rm FedCog}{\rm Update}$ &
		\begin{tabular}[c]{@{}c@{}} Edge Density\\ ($\times10^{-4}$)\end{tabular} \\ \hline
		DBLP         & 1470 & 522  & 2144  & 938  & 0.69 & 1.57 & 3.37  \\
		Coauthor-Phy & 9746 & 2168 & 13183 & 1040 & 0.74 & 9.37 & 4.17  \\
		Flickr       & 4178 & 449  & 1996  & 878  & 2.09 & 4.76 & 83.57 \\
		Amazon-CS    & 1436 & 283  & 825   & 431  & 1.74 & 3.33 & 26.00 \\ \hline
	\end{tabular}
\end{table*}

\bullethdr{Computation Efficiency.}~\cref{tab:eff_compute} shows the run-time of FedCog, FedAvg, and the centralized method during SGC's propagation phase.
For the federated methods (i.e., FedCog and FedAvg), we show the sum of all parties' computing time.
FedAvg is faster than the centralized method because the graph is cut into small pieces and some edges are missing.
Also as expected, FedCog takes more computation resources than FedAvg because FedCog introduces extra nodes and edges to the local graph.
Comparing FedCog with the centralized method, we find that the sum of run-times over all FedCog parties is smaller than the centralized baselines DBLP and Coauthor-Phy datasets, but larger on Flickr and Amazon-CS.
The difference between FedCog and the centralized method in efficiency is caused by two main factors:
(1) Computing on graphs generally costs $O(|V|^2)$ time. Each party $i$ of FedCog works on a small graph with $|V_i|$ nodes, while the centralized method directly computes on a large graph with $|V|$ nodes, so FedCog will benefit from the small local data.
(2) FedCog adds extra nodes and edges to the local graphs, so the total number of nodes and edges is larger than the global graph in the centralized method, which slows down FedCog.
Combining the two factors with the theoretical efficiency analysis in~\cref{sec:efficient} and~\cref{tab:eff_theory}, the number of inter-edges is supposed to be the reason for different efficiency results on different datasets.
The experimental results support our theory.
We list the edge density of each dataset
\begin{equation}
	{\rm density}=\frac{2|E|}{|V|(|V|-1)},
\end{equation}
in~\cref{tab:eff_compute}.
DBLP and Coauthor-Phy have small edge density, thus the number of inter-edges between parties is also small.
FedCog does not need to add too many nodes and edges to the local graphs and presents high efficiency.
On the contrary, Flickr and Amazon-CS have much larger edge densities, so FedCog is slow.
Notice that on the extremely dense graph Flickr, the total computation cost of FedCog is $2.09\times$ that of the centralized method and $0.02\times$ for a single party on average, respectively.
Such computation cost is acceptable in practice.
We also show the computation  cost of federated training in~\cref{tab:eff_compute}.
In the worst case, the computing time of FedCog is approximately $9.37$ rounds of model parameter updates on the Coauthor-Phy dataset.
In our experiments, FedAvg generally needs $100~200$ rounds to reach the best model performance, which means FedCog only consumes at most 10\% of computing resources during the entire federated learning process.

\begin{table}[tbp]
	\centering
	\caption{Communication costs on 100 FL parties}
	\label{tab:eff_communicate}
	\vspace{-10pt}
	\begin{tabular}{c|cc|c}
		\hline
		& \begin{tabular}[c]{@{}c@{}}FedCog\\ (MB)\end{tabular} & \begin{tabular}[c]{@{}c@{}}Global Update\\ (MB/round)\end{tabular} & $\frac{\rm FedCog}{\rm Update}$ \\ \hline
		DBLP         & 8.3    & 125.4 & 0.07 \\
		Coauthor-Phy & 253.5  & 642.5 & 0.39 \\
		Flickr       & 4064.6 & 919.9 & 4.42 \\
		Amazon-CS    & 463.5  & 59.4  & 7.81 \\ \hline
	\end{tabular}
\end{table}

\bullethdr{Communication Efficiency.}~\cref{tab:eff_communicate} shows the communication cost of FedCog.
The FedCog-free or centralized method takes no communication resources during the propagation phase of SGC, so they are not listed in~\cref{tab:eff_communicate}.
We compare FedCog's communication cost with the global parameter updates during federated training.
The worst results show that FedCog's communication cost is equivalent to $7.81$ rounds of global updates on the Amazon-CS dataset, which is a small part of the total communication during federated learning.

\subsection{Results with More Federated Training Algorithms}

\begin{figure}[tbp]
	\centering
	\includegraphics[width=0.97\linewidth]{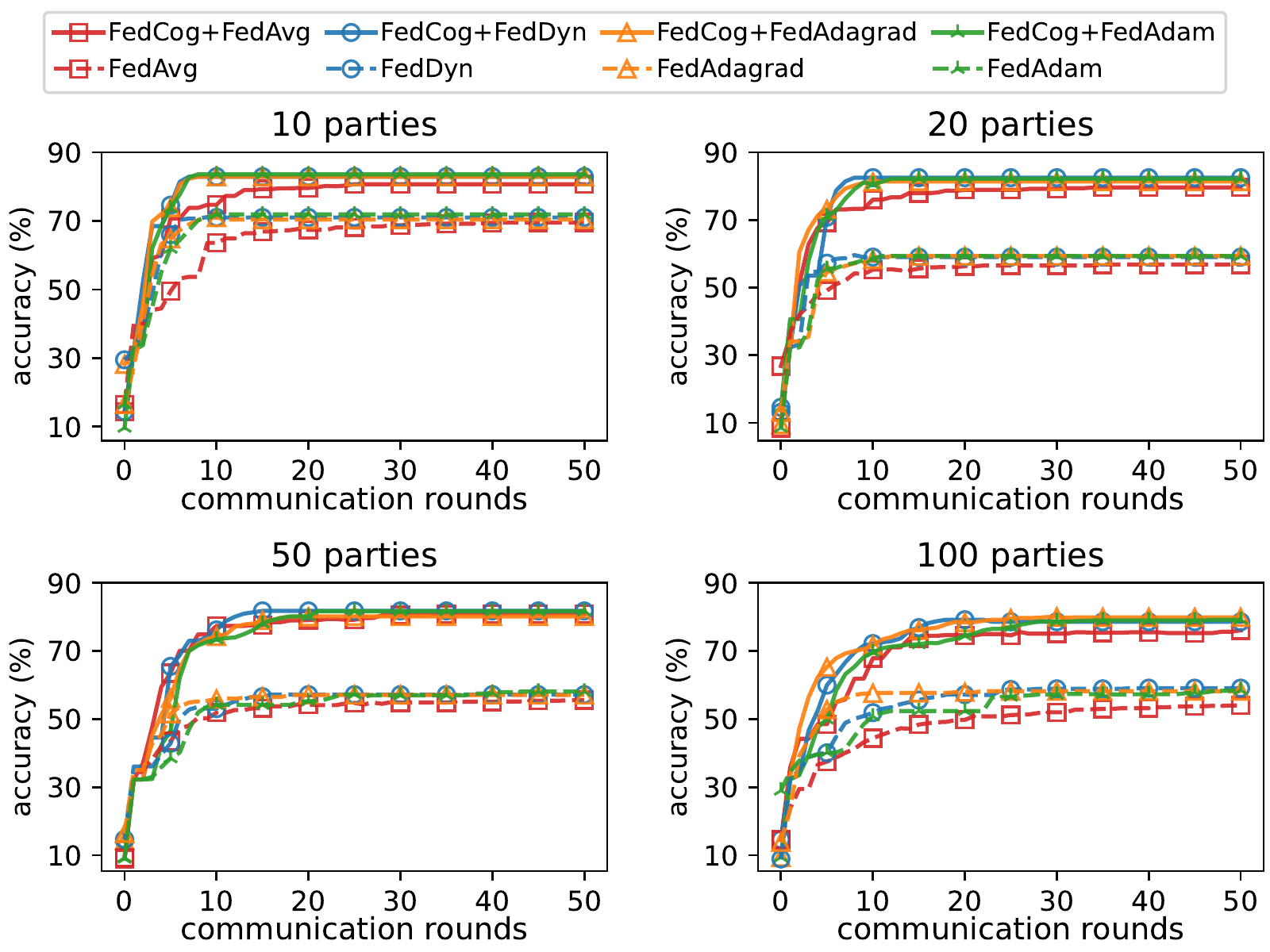}
	\vspace{-10pt}
	\caption{Node classification accuracy on CORA dataset partitioned into $\{10,20,50,100\}$ parts with K-Means. Solid lines show the results of FedCog, and dashed lines are for SGC without regard to coupled graphs.}
\label{fig:fed_alg}
\end{figure}

All model parameters are trained by FedAvg~\cite{McMahanMRHA17} (i.e., \cref{alg:fedcog_sgc}) during experiments in \cref{sec:res_nc,sec:res_lp,sec:res_eff}.
In this section, we show the results of training the SGC model with FedDyn~\cite{AcarZNMWS21}, FedAdagrad, and FedAdam~\cite{ReddiCZGRKKM21}, which are designed to improve FL convergence, especially for non-IID local data.
\cref{fig:fed_alg} shows the node classification accuracy over the CORA dataset during the first 50 communication rounds of federated training.
FedDyn, FedAdagrad, and FedAdam with FedCog (resp. without FedCog) provide better convergence rates and model performance than FedAvg with FedCog (resp. without FedCog).
However, the improved FL training algorithms cannot fill the gaps between performance with and without FedCog.
For example, the accuracy of FedAvg is 54.0\% at the 50-th round in the 100-party scenario.
The results of FedAdagrad and FedCog+FedAvg are 59.1\% and 76.1\%, respectively.
Such results suggest that the coupled graph problem is not merely caused by local data's non-IID distribution.
The lack of topological information in graph federated learning cannot be offset by simply enhancing the performance of model parameter update algorithms.
Methods for dealing with inter-party graph data (such as FedCog) are essential for graph federated learning.

\section{Related Work}\label{sec:related}

\header{Federated Learning} (FL)~\cite{McMahanMRHA17,YangLCT19} is a machine learning framework where several parties (or, clients) jointly train a model while preserving the data privacy of each party.
There are three classes of FL based on the partition pattern of data, i.e., horizontal FL, vertical FL, and transfer FL.
Horizontal FL is designed for the situation where each party shares the same attribute and label space, while the data samples are partitioned.
Vertical FL is the opposite of horizontal FL and it considers that parties have the same samples but the attribute or label spaces are different.
Transfer FL allows the parties to have few samples or attributes and uses the transfer learning methods to build a federated model.
Recent works in the FL area concentrate on communication efficiency~\cite{ReisizadehMHJP20,RothchildPUISB020,ZhouYL22,TangNW00L022}, 
privacy preservation~\cite{WuKLHFPY22,ZhangCH0Y22,ChenOK22,GongSKWCDI22}, 
and robustness on not independent and identically distributed (Non-IID) data~\cite{HsuQ020,AcarZNMWS21,ReddiCZGRKKM21,ZhangS0TD22}.

\header{Graph Neural Networks} (GNN)~\cite{1555942,ScarselliGTHM09} are deep learning methods to solve the graph learning tasks such as node classification and link prediction.
Graph Convolutional Network (GCN)~\cite{KipfW17} is a GNN model with graph convolution operators.
From the topological perspective, GCN can be interpreted as feature propagation operations, i.e., each node aggregates the features of its neighbors.
SGC~\cite{WuSZFYW19} adopts straightforward low-pass filters to simplify GCN,
which smooths the neighbor features using normalized adjacency matrices.
GNN models, especially GCN and its variants, have been successfully applied to recommendation systems~\cite{YingHCEHL18,WuZGBC20,ZhengGCJL21}, 
social networks mining~\cite{WuLXWC20,LinGL20}, 
natural language processing~\cite{MarcheggianiT17,LiG19,DongNYL22}, 
and biochemistry~\cite{FoutBSB17}.

The most related line of research to our work is the \textbf{FL over Graphs}.
It is non-trivial to perform FL over graphs because graphs are non-Euclidean data and particularly they are given in a variety of forms for real-world applications.
To address this challenge, recently, several methods have been proposed and they can be summarized as follows.

\textbullet\, The \emph{horizontal graph FL} mainly has three classes, i.e., {\em graph level} FL, {\em subgraph level} FL, and {\em node level} FL.
For graph level FL, each party has a set of graphs.
Take FedGraphNN~\cite{abs-2104-07145} as an instance.
FedGraphNN exploits small graphs (e.g., chemical and biological molecular graphs) distributed on multiple parties and uses FL to achieve high performance of graph classification and regression.
Subgraph level FL considers each party as a collection of nodes and edges from a graph.
For example, each mobile user holds a subgraph consisting of edges to its friends or its installed mobile applications.
Mobile users can jointly perform FL on their personal data to build an effective system for recommending mobile applications.
To the best of our knowledge, FedGNN~\cite{abs-2102-04925} is the first subgraph level FL.
GraphFL~\cite{abs-2012-04187} provides a generalized method for subgraph level FL, and ASFGNN~\cite{ZhengZCWWZ21} addresses the problem of non-IID node features in subgraph level FL.
FedGraphNN~\cite{abs-2104-07145} also discusses the settings of partitioned subgraphs and nodes, but it does not involve the data coupling among parties, and only fits some specific cases (i.e., it is mainly designed for bipartite graphs).
For node level FL, each party processes only one node and all of its connections in the graph of interest.
The practical cases include online social networks and spatial-temporal trajectory graphs~\cite{abs-2106-05223}.

\textbullet\, The \emph{vertical graph FL} studies different node data partition scenarios.
VFGNN~\cite{abs-2005-11903} focuses on the scenario where node features are vertically separated into different parties.
SGNN~\cite{MeiGLP19} performs the vertical graph FL when each party has all nodes in the graph but partial content, structural and label information.
FedGL~\cite{abs-2105-03170} further considers the situation where the nodes on parties are not completely overlapped.

In the previous graph FL researches, the partition of node features and edge features are studied.
Horizontal graph FL regards the graph data on parties as individual graphs or topologically independent subgraphs.
Vertical graph FL focuses on the division of attribute space, where the nodes and edges are shared among parties.
Therefore, a majority of previous works on graph FL do not consider the partition of graph topological structures.
Topological structures are essential components for graphs and play important roles in graph learning.
In our work, we consider not only the partition of node samples, but also the partition of the graph's topological structures.
We propose \emph{coupled graphs} to represent the distributed graph data.
A recent work, FedGCN~\cite{abs-2201-12433}, also focuses on a similar problem as our work.
It addresses the cross-party edge problem in graph federated learning.
In FedGCN, each federated learning party receives node features from other parties and stores them on its local device.
Thus, the locally stored graph data from other parties can participate in local computation.
However, it faces some challenges.
The direct transmission of node features lacks privacy guarantees.
There are risks of sensitive data leakage, e.g., edges and node features.
In addition, the communication and storage costs of FedGCN are not minor, especially influenced by the depth of the GNN model.
The expensive cost of multi-layer models is also a problem of FedGCN.
In our work, we propose a graph decoupling method to deal with the edges between federated parties.
Our method also provides a guarantee for the privacy of node features and has practical communication costs.

\section{Conclusion} \label{sec:conclusions}
In this work, we formally define the coupled graph learning problem in the setting of Federated Learning, and propose a framework FedCog to address it.
To improve the performance of federated models,
FedCog enables the federated parties to share the coupled information in a privacy-preserving way.
The experiments demonstrate that our FedCog increases the performance of federated graph learning by up to $14.7\%$ accuracy in node classification tasks compared with existing baselines, and up to $0.232$ AUC score in link prediction tasks.
In the future, we will work towards communication efficiency and privacy preservation strategies for more GNN models.

\ifCLASSOPTIONcompsoc
  \section*{Acknowledgments}
\else
  \section*{Acknowledgment}
\fi
The authors would like to thank the anonymous reviewers for their comments and suggestions.
This work was supported in part by 
National Key R\&D Program of China (2021YFB1715600), 
National Natural Science Foundation of China (U22B2019, 62272372, 61902305), 
MoE-CMCC "Artificial Intelligence" Project (MCM20190701).

\bibliographystyle{unsrt}
\bibliography{references}

\end{document}